\documentclass{article} % For LaTeX2e
\usepackage{iclr2025_conference,times}

\usepackage{amsmath,amsfonts,bm}

\def\eqref#1{equation~\ref{#1}}
\def\1{\bm{1}}

\DeclareMathAlphabet{\mathsfit}{\encodingdefault}{\sfdefault}{m}{sl}
\SetMathAlphabet{\mathsfit}{bold}{\encodingdefault}{\sfdefault}{bx}{n}

\usepackage{hyperref}
\usepackage{url}
\usepackage{booktabs} 
\usepackage{graphicx}

\usepackage{booktabs}
\usepackage{multirow}
\usepackage{colortbl}
\usepackage{pifont}

\usepackage{tabularx}
\usepackage{pdflscape}

\usepackage{array}           % For better column control
\usepackage{siunitx}         % For numerical alignment
\usepackage{caption}         % For better caption control
\usepackage{colortbl}        % For cell coloring
\usepackage{makecell}        % For multi-line headers (optional)
\usepackage{pifont}% http://ctan.org/pkg/pifont
\usepackage{xcolor}
\usepackage{pgf}         % For calculations and color mixing
\usepackage{microtype}
\definecolor{highlight}{gray}{0.90}

\definecolor{good}{rgb}{0.24, 0.7, 0.44}
\definecolor{headergray}{gray}{0.9}    % Light gray for header
\definecolor{rowgray}{gray}{0.95}      % Very light gray for alternating rows
\definecolor{neutral}{gray}{0.6}        % Gray for neutral improvements

\title{HVT: A Comprehensive Vision Framework for Learning in Non-Euclidean Space}

\iclrfinalcopy
\author{%
  Jacob Fein-Ashley \\
  University of Southern California\\
  \texttt{feinashl@usc.edu} \\
  \And
  Ethan Feng \\
  University of Southern California \\
  \texttt{eyfeng@usc.edu} \\
  \AND
  Minh Pham \\
  University of Southern California \\
  \texttt{tranppha@usc.edu} \\
}

\begin{document}

\maketitle

\begin{abstract}
    Data representation in non-Euclidean spaces has proven effective for capturing hierarchical and complex relationships in real-world datasets. Hyperbolic spaces, in particular, provide efficient embeddings for hierarchical structures. This paper introduces the Hyperbolic Vision Transformer (HVT), a novel extension of the Vision Transformer (ViT) that integrates hyperbolic geometry. While traditional ViTs operate in Euclidean space, our method enhances the self-attention mechanism by leveraging hyperbolic distance and Möbius transformations. This enables more effective modeling of hierarchical and relational dependencies in image data. We present rigorous mathematical formulations, showing how hyperbolic geometry can be incorporated into attention layers, feed-forward networks, and optimization. We offer improved performance for image classification using the ImageNet dataset.
\end{abstract}

\section{Introduction}
Representation learning is fundamental to modern machine learning, enabling models to extract meaningful features from raw data \cite{bengio2013representation}. While Euclidean spaces are traditionally used to model data relationships, many real-world datasets---including images---exhibit hierarchical structures better captured in non-Euclidean spaces \cite{bronstein2017geometric}.

Images are inherently hierarchical, comprising structures at multiple scales: from pixels to edges, from shapes to objects, and ultimately to entire scenes \cite{biederman1987recognition,riesenhuber1999hierarchical}. This hierarchy can be conceptualized as:

\begin{itemize}
    \item \textbf{Pixels} $\{p_i\}$: The basic units of an image.
    \item \textbf{Edges} $\{e_j\}$: Formed by grouping pixels with significant intensity gradients.
    \item \textbf{Shapes} $\{s_k\}$: Created by combining edges into simple motifs.
    \item \textbf{Objects} $\{o_l\}$: Recognizable entities composed of shapes.
    \item \textbf{Scenes} $\mathcal{I}$: Complete images where objects interact within a context.
\end{itemize}

The hierarchical nature of images implies that higher-level concepts are built upon lower-level features, reflecting a tree-like structure. Vision Transformers (ViTs) \cite{dosovitskiy2020image} process images by dividing them into patches, treating each patch as a token. This patch-based approach introduces a hierarchical representation because:

\begin{enumerate}
    \item \textbf{Local Features}: Each patch captures local patterns such as textures or edges.
    \item \textbf{Global Context}: By attending over patches, the model aggregates local information to understand the overall structure.
\end{enumerate}

This mirrors the hierarchical composition of images, from local to global features.

Hyperbolic spaces are well-suited for modeling hierarchical data due to their ability to embed tree-like structures with minimal distortion \cite{nickel2017poincare}. By utilizing hyperbolic geometry and M\"obius transformations, we can effectively capture the multi-scale dependencies inherent in visual data \cite{ganea2018hyperbolic}. Specifically, M\"obius transformations allow for operations like addition and scalar multiplication in hyperbolic space, enabling neural networks to perform calculations while preserving hierarchical relationships.

In this paper, we propose the \emph{Hyperbolic Vision Transformer}, which integrates hyperbolic geometry into the transformer architecture. Our contributions include:

\begin{itemize}
    \item \textbf{Hyperbolic Neural Components}: Extending ViT to operate in hyperbolic space using hyperbolic versions of neural network components, such as attention mechanisms and linear layers.
    \item \textbf{M\"obius Transformations in ViT}: Demonstrating how M\"obius transformations enable operations in hyperbolic space, preserving hierarchical data structures.
    \item \textbf{Theoretical and Empirical Analysis}: Providing insights and evaluations showing improved modeling of hierarchical structures over traditional Euclidean approaches.
\end{itemize}

\section{Related Work}
\subsection{Hyperbolic Geometry in Machine Learning}

Recent advances have leveraged hyperbolic geometry for machine learning, significantly impacting how hierarchical data is modeled. \citet{nickel2017poincare} successfully used Poincaré embeddings for such data, showing marked improvements over Euclidean embeddings. This approach was further refined by \citet{ganea2018hyperbolic}, who introduced hyperbolic embeddings for entailment cones, effectively capturing asymmetric relationships.

\citet{khrulkov2020hyperbolic} and \citet{liu2020hyperbolic} extended these concepts to visual data and zero-shot recognition, respectively, underscoring the versatility of hyperbolic embeddings in handling complex visual tasks. Most notably, \citet{ermolov2022hyperbolic} developed Hyperbolic Vision Transformers (HVTs) that incorporate these embeddings within Vision Transformer architectures, enhancing metric learning. Our model extends this integration by incorporating hyperbolic geometry throughout the transformer operations, from Möbius transformations to hyperbolic self-attention.

\subsection{Hyperbolic Neural Networks and Attention}

\citet{ganea2018hyperbolic} initially formulated hyperbolic neural networks, introducing layers and activation functions suited for hyperbolic spaces. This foundation was expanded by \citet{bachmann2020constant}, who focused on optimizing these networks efficiently. Our method enriches this foundation by embedding hyperbolic layers directly within transformer architectures, enhancing the adaptability and depth of hyperbolic operations.

Another method presented in Hyperbolic Attention Networks~\cite{gulcehre2018hyperbolicattentionnetworks} differs from ours primarily in how hyperbolic geometry is applied within the attention mechanism. Gulcehre et al.\ focus on embedding the activations into hyperbolic space using both the hyperboloid and Klein models, leveraging hyperbolic matching and aggregation operations. In contrast, our approach incorporates learnable curvature within positional embeddings, head-specific scaling in attention, and hyperbolic layer normalization, offering more flexibility and efficiency in capturing hierarchical data. Additionally, we utilize the Poincaré ball model for its computational suitability in vision tasks.

\subsection{Vision Transformers}

Initially developed for NLP by \citet{vaswani2017attention}, Vision Transformers (ViT) have been adapted for visual tasks, as demonstrated by \citet{dosovitskiy2021image}. Enhancements such as those proposed by \citet{caron2021emerging} and \citet{el2021training} have refined ViTs for self-supervised learning. Unlike these Euclidean-based models, our Hyperbolic Vision Transformer employs hyperbolic geometry to model hierarchical and relational data structures more effectively.

\subsection{Comparison to Key Hyperbolic Methods}
While \citet{ermolov2022hyperbolic} introduced HVTs that focus on hyperbolic embeddings for metric learning, our approach fully integrates hyperbolic operations throughout the transformer, significantly enhancing its ability to manage hierarchical data beyond the objectives of metric learning. We extend the application scope to include image classification by embedding hyperbolic geometry directly into the core components of the Vision Transformer.

Recently, \citet{yang2024hypformer} proposed Hypformer, an efficient hyperbolic Transformer based on the Lorentz model of hyperbolic geometry. Hypformer introduces two foundational blocks—Hyperbolic Transformation with Curvatures (HTC) and Hyperbolic Readjustment and Refinement with Curvatures (HRC)—to define essential Transformer modules in hyperbolic space. They also develop a linear self-attention mechanism in hyperbolic space to handle large-scale graph data and long-sequence inputs efficiently.

While Hypformer makes significant contributions to the development of hyperbolic Transformers, particularly in processing large-scale graph data, our model differs in several key aspects:

\begin{itemize}
    \item \textbf{Model Focus and Application Domain}: Hypformer is primarily designed for graph data and emphasizes scalability and efficiency in handling large-scale graphs and long sequences. In contrast, our model focuses on vision tasks, specifically image classification, integrating hyperbolic geometry throughout the Vision Transformer architecture to enhance the modeling of hierarchical and relational structures inherent in visual data.

    \item \textbf{Hyperbolic Model Used}: Hypformer operates in the Lorentz model of hyperbolic geometry, whereas we utilize the Poincaré ball model. The Poincaré ball model is advantageous for vision tasks due to its conformal properties, which preserve angles and better represent geometric structures in image data.

    \item \textbf{Innovative Components}: Our model introduces unique components such as learnable curvature in positional embeddings, head-specific scaling in the attention mechanism, hyperbolic layer normalization, gradient clipping, geodesic regularization, and layer scaling for training stability. These innovations are specifically tailored to enhance the performance of Vision Transformers in hyperbolic space.

    \item \textbf{Attention Mechanism}: While Hypformer develops a linear self-attention mechanism to address efficiency in handling large-scale data, our model extends the standard self-attention mechanism into hyperbolic space using Möbius operations and hyperbolic distance calculations. This approach allows us to capture complex relationships in visual data more effectively.

\end{itemize}

By fully integrating hyperbolic operations within the Vision Transformer framework and focusing on the unique challenges of vision tasks, our model offers a comprehensive solution that provides superior performance over previous hyperbolic Transformers, including Hypformer, in the domain of image classification.

\section{Proposed Method}
This section presents the mathematical foundations of the Hyperbolic Vision Transformer Network (HVT). We cover essential concepts from hyperbolic geometry and describe how we adapt Vision Transformer components to operate in hyperbolic space. Our main contributions include introducing learnable curvature in positional embeddings, head-specific scaling in attention mechanisms, hyperbolic layer normalization, gradient clipping, geodesic regularization, and layer scaling for training stability.

\scriptsize{\textbf{Code Availability}: Code is available at \url{https://github.com/hyperbolicvit/hyperbolicvit}}
\normalsize

\subsection{Hyperbolic Geometry Preliminaries}

Hyperbolic space, characterized by constant negative curvature, embeds hierarchical and complex structures common in visual data. We adopt the Poincaré ball model due to its computational convenience and suitability for representing image data structures.

 \subsubsection{Poincaré Ball Model}

The \( n \)-dimensional Poincaré ball model is defined as the manifold:
\begin{equation}
    \mathbb{D}^n = \left\{ \mathbf{x} \in \mathbb{R}^n : \|\mathbf{x}\| < 1 \right\},
\end{equation}
where \( \|\cdot\| \) denotes the Euclidean norm. The Riemannian metric tensor \( g_{\mathbf{x}} \) of this manifold is given by:
\begin{equation}
    g_{\mathbf{x}} = \lambda_{\mathbf{x}}^2 g^E, \quad \text{with} \quad \lambda_{\mathbf{x}} = \frac{2}{1 - \|\mathbf{x}\|^2},
    \label{eq:riemannian_metric}
\end{equation}
where \( g^E \) is the Euclidean metric tensor, and \( \lambda_{\mathbf{x}} \) is the conformal factor that scales the Euclidean metric to account for the curvature of hyperbolic space.

\subsubsection{Möbius Operations}

To adapt Vision Transformer components to hyperbolic space, we utilize Möbius transformations, essential for processing vectors within the Poincaré ball model.

\paragraph{Möbius Addition}

For vectors \( \mathbf{x}, \mathbf{y} \in \mathbb{D}^n \), the Möbius addition \( \mathbf{x} \oplus \mathbf{y} \) is defined as:
\begin{equation}
    \mathbf{x} \oplus \mathbf{y} = \frac{(1 + 2 \langle \mathbf{x}, \mathbf{y} \rangle + \|\mathbf{y}\|^2)\mathbf{x} + (1 - \|\mathbf{x}\|^2)\mathbf{y}}{1 + 2 \langle \mathbf{x}, \mathbf{y} \rangle + \|\mathbf{x}\|^2 \|\mathbf{y}\|^2},
    \label{eq:mobius_addition}
\end{equation}
where \( \langle \cdot , \cdot \rangle \) denotes the Euclidean inner product. Möbius addition generalizes vector addition to hyperbolic space, ensuring the result remains within the manifold.

\paragraph{Möbius Scalar Multiplication}

For a scalar \( r \in \mathbb{R} \) and a vector \( \mathbf{x} \in \mathbb{D}^n \), Möbius scalar multiplication \( r \otimes \mathbf{x} \) is defined as:
\begin{equation}
    r \otimes \mathbf{x} = \tanh\left( r \tanh^{-1}(\|\mathbf{x}\|) \right) \frac{\mathbf{x}}{\|\mathbf{x}\|}.
    \label{eq:mobius_scalar_multiplication}
\end{equation}
This operation scales a vector while preserving its direction and ensures the scaled vector remains within the Poincaré ball.

\paragraph{Möbius Matrix-Vector Multiplication}

Given a matrix \( \mathbf{W} \in \mathbb{R}^{m \times n} \) and a vector \( \mathbf{x} \in \mathbb{D}^n \), Möbius matrix-vector multiplication \( \mathbf{W} \otimes_M \mathbf{x} \) is defined as:
\begin{equation}
    \mathbf{W} \otimes_M \mathbf{x} = \tanh \left( \frac{\|\mathbf{W} \mathbf{x}\|}{\|\mathbf{x}\|} \tanh^{-1}(\|\mathbf{x}\|) \right) \frac{\mathbf{W} \mathbf{x}}{\|\mathbf{W} \mathbf{x}\|}, \quad \text{if} \quad \mathbf{x} \ne \mathbf{0}.
    \label{eq:mobius_matrix_vector_multiplication}
\end{equation}
This operation extends Möbius scalar multiplication to linear transformations, allowing us to apply linear layers within hyperbolic space.

\paragraph{Möbius Concatenation}

To combine multiple vectors \( \mathbf{x}_1, \mathbf{x}_2, \ldots, \mathbf{x}_n \in \mathbb{D}^n \), we use Möbius concatenation:
\begin{equation}
    \bigoplus_{j=1}^n \mathbf{x}_j = \mathbf{x}_1 \oplus \mathbf{x}_2 \oplus \cdots \oplus \mathbf{x}_n.
\end{equation}

\subsection{Hyperbolic Neural Network Components}

We incorporate the hyperbolic operations into our neural network components to enable the Vision Transformer architecture to operate within hyperbolic space.

\subsubsection{Hyperbolic Linear Layer}

Traditional linear layers are adapted to hyperbolic space using Möbius matrix-vector multiplication followed by Möbius addition with a bias:
\begin{equation}
    \mathbf{h} = \mathbf{W} \otimes_M \mathbf{x} \oplus \mathbf{b},
    \label{eq:hyperbolic_linear_layer}
\end{equation}
where \( \mathbf{x} \in \mathbb{D}^{n} \), \( \mathbf{W} \in \mathbb{R}^{m \times n} \), and \( \mathbf{b} \in \mathbb{D}^{m} \). This layer allows us to perform linear transformations while respecting the geometry of hyperbolic space.

\subsubsection{Hyperbolic Activation and Normalization}

Activation functions and normalization are applied in the tangent space at the origin to leverage familiar Euclidean operations. The logarithmic map \( \log_0: \mathbb{D}^n \rightarrow T_0\mathbb{D}^n \) maps points from the manifold to the tangent space:
\begin{equation}
    \log_0(\mathbf{x}) = \frac{2 \tanh^{-1}(\|\mathbf{x}\|)}{\|\mathbf{x}\|} \mathbf{x}.
    \label{eq:log_map}
\end{equation}
The exponential map \( \exp_0: T_0\mathbb{D}^n \rightarrow \mathbb{D}^n \) brings points back to the manifold:
\begin{equation}
    \exp_0(\mathbf{v}) = \tanh\left( \frac{\|\mathbf{v}\|}{2} \right) \frac{\mathbf{v}}{\|\mathbf{v}\|}.
    \label{eq:exp_map}
\end{equation}
Using these maps, we define hyperbolic versions of the ReLU activation and Layer Normalization:
\begin{align}
    \text{ReLU}^\mathbb{D}(\mathbf{x}) &= \exp_0 \left( \text{ReLU} \left( \log_0(\mathbf{x}) \right) \right), \label{eq:hyperbolic_relu} \\
    \text{LayerNorm}^\mathbb{D}(\mathbf{x}) &= \exp_0 \left( \text{LayerNorm} \left( \log_0(\mathbf{x}) \right) \right). \label{eq:hyperbolic_layernorm}
\end{align}
This approach allows us to apply standard activation and normalization techniques within hyperbolic space.

\subsubsection{Hyperbolic Layer Scaling}

To stabilize residual connections in hyperbolic space, we introduce a learnable scaling parameter \( \beta \):
\begin{equation}
    \mathbf{X}' = \mathbf{X} \oplus \beta \otimes \mathbf{O},
    \label{eq:hyperbolic_layer_scaling}
\end{equation}
where \( \mathbf{X} \) is the input, \( \mathbf{O} \) is the output from a layer, and \( \beta \in \mathbb{R} \) scales the residual contribution.

\subsection{Hyperbolic Vision Transformer Architecture}

We integrate hyperbolic geometry into the Vision Transformer architecture by modifying key components to operate within hyperbolic space.

\subsubsection{Learnable Hyperbolic Positional Embeddings}

Positional embeddings are adjusted using a learnable curvature parameter \( c \) to represent positional information in hyperbolic space better:
\begin{align}
    \mathbf{E}_{\text{pos}} &= c \otimes \mathbf{E}_{\text{pos}}^0, \\
    \mathbf{X} &= \mathbf{X} \oplus \mathbf{E}_{\text{pos}},
    \label{eq:hyperbolic_positional_embeddings}
\end{align}
where \( \mathbf{E}_{\text{pos}}^0 \in \mathbb{D}^{1 \times N \times E} \) are the initial positional embeddings.

\subsubsection{Hyperbolic Self-Attention Mechanism}

We extend the self-attention mechanism to hyperbolic space using Möbius operations and hyperbolic distance computations. Table~\ref{tab:symbols} summarizes the symbols used.

\begin{table}[ht]
\centering
\caption{Symbols and their descriptions.}
\label{tab:symbols}

% Reduce font size to fit content
\scriptsize

% Create the table using tabularx for adjustable-width columns
\begin{tabularx}{\columnwidth}{@{} >{\bfseries}l X @{}}
\toprule
\rowcolor{headergray}
\textbf{Symbol} & \textbf{Description} \\
\midrule
\rowcolor{white}
\( B \)        & Batch size. \\
\rowcolor{rowgray}
\( H \)        & Number of attention heads. \\
\rowcolor{white}
\( N \)        & Sequence length. \\
\rowcolor{rowgray}
\( D \)        & Head dimension, where \( D = \frac{E}{H} \) and \( E \) is the embedding dimension. \\
\rowcolor{white}
\( c \)        & Curvature parameter of the Poincaré ball model. \\
\rowcolor{rowgray}
\( \epsilon \) & Small constant for numerical stability \( (\epsilon = 1 \times 10^{-15}) \). \\
\rowcolor{white}
\( \delta \)   & Small constant for clamping \( (\delta = 1 \times 10^{-7}) \). \\
\rowcolor{rowgray}
\( p \)        & DropConnect~\cite{dropconnect} probability. \\
\bottomrule
\end{tabularx}
\end{table}

Given input embeddings \( \mathbf{X} \in \mathbb{D}^{B \times N \times E} \), we compute the queries \( \mathbf{Q} \), keys \( \mathbf{K} \), and values \( \mathbf{V} \) using hyperbolic linear layers:

\begin{align}
    \mathbf{Q} &= \mathbf{X} \otimes_M \mathbf{W}_Q \oplus \mathbf{b}_Q, \\
    \mathbf{K} &= \mathbf{X} \otimes_M \mathbf{W}_K \oplus \mathbf{b}_K, \\
    \mathbf{V} &= \mathbf{X} \otimes_M \mathbf{W}_V \oplus \mathbf{b}_V.
\end{align}
We reshape these tensors to accommodate multiple attention heads:
\begin{equation}
    \mathbf{Q}, \mathbf{K}, \mathbf{V} \in \mathbb{D}^{B \times H \times N \times D}.
\end{equation}

\paragraph{Hyperbolic Distance Computation}

The hyperbolic distance between a query \( \mathbf{Q}_{b,h,i} \) and a key \( \mathbf{K}_{b,h,j} \) is computed using the distance function in the Poincaré ball model:
\begin{equation}
    d_\mathbb{D}(\mathbf{Q}_{b,h,i}, \mathbf{K}_{b,h,j}) = \cosh^{-1}\left( 1 + \frac{2c \|\mathbf{Q}_{b,h,i} \oplus (-\mathbf{K}_{b,h,j})\|^2}{(1 - c \|\mathbf{Q}_{b,h,i}\|^2)(1 - c \|\mathbf{K}_{b,h,j}\|^2) + \epsilon} \right).
    \label{eq:hyperbolic_distance}
\end{equation}
To ensure numerical stability, we clamp the argument of \( \cosh^{-1} \):
\begin{equation}
    \mathcal{A}_{b,h,i,j} = \max\left(1 + \frac{2c \|\mathbf{Q}_{b,h,i} \oplus (-\mathbf{K}_{b,h,j})\|^2}{(1 - c \|\mathbf{Q}_{b,h,i}\|^2)(1 - c \|\mathbf{K}_{b,h,j}\|^2) + \epsilon},\ 1 + \delta\right).
    \label{eq:clamped_argument}
\end{equation}
We then compute the distance as:
\begin{equation}
    d_\mathbb{D}(\mathbf{Q}_{b,h,i}, \mathbf{K}_{b,h,j}) = \log\left( \mathcal{A}_{b,h,i,j} + \sqrt{\mathcal{A}_{b,h,i,j}^2 - 1} \right).
    \label{eq:hyperbolic_distance_final}
\end{equation}

\paragraph{Attention Scores and Weights}

Using the computed distances, we calculate the attention scores with a head-specific scaling factor \( \alpha_h \):
\begin{equation}
    \mathbf{A}_{b,h,i,j} = -\frac{d_\mathbb{D}(\mathbf{Q}_{b,h,i}, \mathbf{K}_{b,h,j})^2}{\alpha_h \sqrt{D}}.
    \label{eq:attention_scores}
\end{equation}
The attention weights are obtained by normalizing the scores using the softmax function:
\begin{equation}
    \alpha_{b,h,i,j} = \frac{\exp(\mathbf{A}_{b,h,i,j})}{\sum_{k=1}^{N} \exp(\mathbf{A}_{b,h,i,k})}.
    \label{eq:attention_weights}
\end{equation}

\paragraph{Output Computation}

Each attention head produces an output by aggregating the value vectors, weighted by the attention weights:
\begin{equation}
    \mathbf{O}_{b,h,i} = \bigoplus_{j=1}^{N} \left( \alpha_{b,h,i,j} \otimes \mathbf{V}_{b,h,j} \right).
    \label{eq:head_output}
\end{equation}
The outputs from all heads are combined using Möbius concatenation and transformed with a final linear layer:
\begin{equation}
    \mathbf{O}_{b,i} = \mathbf{W}_O \otimes_M \left( \bigoplus_{h=1}^{H} \mathbf{O}_{b,h,i} \right) \oplus \mathbf{b}_O.
    \label{eq:final_output}
\end{equation}
DropConnect regularization is applied to the attention weights during training with probability \( p \).

\paragraph{Residual Connection with Layer Scaling}

As before, we use a residual connection with a learnable scaling parameter \( \beta \):
\begin{equation}
    \mathbf{X}' = \mathbf{X} \oplus \beta \otimes \mathbf{O}.
    \label{eq:attention_residual}
\end{equation}

\subsubsection{Hyperbolic Feed-Forward Network}

The feed-forward network in hyperbolic space consists of two hyperbolic linear layers with a hyperbolic ReLU activation in between:
\begin{align}
    \mathbf{H}_1 &= \mathbf{X}' \otimes_M \mathbf{W}_1 \oplus \mathbf{b}_1, \label{eq:ffn_layer1} \\
    \mathbf{H}_2 &= \text{ReLU}^\mathbb{D}(\mathbf{H}_1), \label{eq:ffn_activation} \\
    \mathbf{H}_3 &= \mathbf{H}_2 \otimes_M \mathbf{W}_2 \oplus \mathbf{b}_2, \label{eq:ffn_layer2}
\end{align}
followed by a residual connection and hyperbolic layer normalization:
\begin{equation}
    \mathbf{X}'' = \text{LayerNorm}^\mathbb{D} \left( \mathbf{X}' \oplus \beta \otimes \mathbf{H}_3 \right).
    \label{eq:ffn_residual}
\end{equation}

\subsection{Optimization in Hyperbolic Space}

Optimizing neural networks in hyperbolic space presents unique challenges due to its non-Euclidean nature. We employ several techniques to ensure stable and effective training.

\paragraph{Gradient Clipping}

To prevent large gradients from destabilizing training, we clip gradients in the tangent space:
\begin{equation}
    \operatorname{grad}_{\text{clipped}} = 
    \begin{cases}
        \operatorname{grad}_{\boldsymbol{\theta}} & \text{if } \|\operatorname{grad}_{\boldsymbol{\theta}}\| \leq v, \\
        v \cdot \dfrac{\operatorname{grad}_{\boldsymbol{\theta}}}{\|\operatorname{grad}_{\boldsymbol{\theta}}\|} & \text{otherwise},
    \end{cases}
    \label{eq:gradient_clipping}
\end{equation}
where \( \|\cdot\| \) is the Euclidean norm in the tangent space, and \( v \) is the clipping threshold.

\paragraph{Riemannian Adam Optimizer}

We utilize the Riemannian Adam optimizer, which adapts the Adam optimization algorithm to Riemannian manifolds. The update rule is:
\begin{equation}
    \boldsymbol{\theta}_{t+1} = \exp_{\boldsymbol{\theta}_t} \left( -\eta_t \cdot \frac{m_t^{\text{clipped}}}{\sqrt{v_t^{\text{clipped}}} + \epsilon} \right),
    \label{eq:riemannian_adam}
\end{equation}
where \( \eta_t \) is the learning rate, \( m_t^{\text{clipped}} \) and \( v_t^{\text{clipped}} \) are the clipped first and second moment estimates, and \( \exp_{\boldsymbol{\theta}_t} \) is the exponential map at \( \boldsymbol{\theta}_t \).

\paragraph{Geodesic Distance Regularization}

To enhance class separation and encourage meaningful representations, we introduce a regularization term based on geodesic distances:
\begin{equation}
    \mathcal{L}_{\text{geo}} = \lambda_{\text{reg}} \cdot \mathbb{E}_{(i,j)} \left[ (1 - \delta_{y_i y_j}) \cdot d_\mathbb{D}(\mathbf{x}_i, \mathbf{x}_j) \right],
    \label{eq:geodesic_regularization}
\end{equation}
where \( \delta_{y_i y_j} \) is the Kronecker delta function indicating whether samples \( i \) and \( j \) belong to the same class. The total loss objective is \(\mathcal{L}_{\text{cross entropy}} + \mathcal{L}_{\text{geo}}\).

\paragraph{Layer Scaling and Attention Scaling}

Introducing the learnable scaling parameter \( \beta \) in residual connections and head-specific scaling factors \( \alpha_h \) in the attention mechanism helps control the magnitude of updates. It allows each attention head to adapt its sensitivity to hyperbolic distances.

\paragraph{Parameter Initialization}

Weights are initialized using Xavier uniform initialization adapted for hyperbolic space~\cite{leimeister2018skip}. All manifold parameters are initialized within the Poincaré ball to ensure valid representations and stable training.

\begin{figure}
    \centering
    \includegraphics[scale=.4]{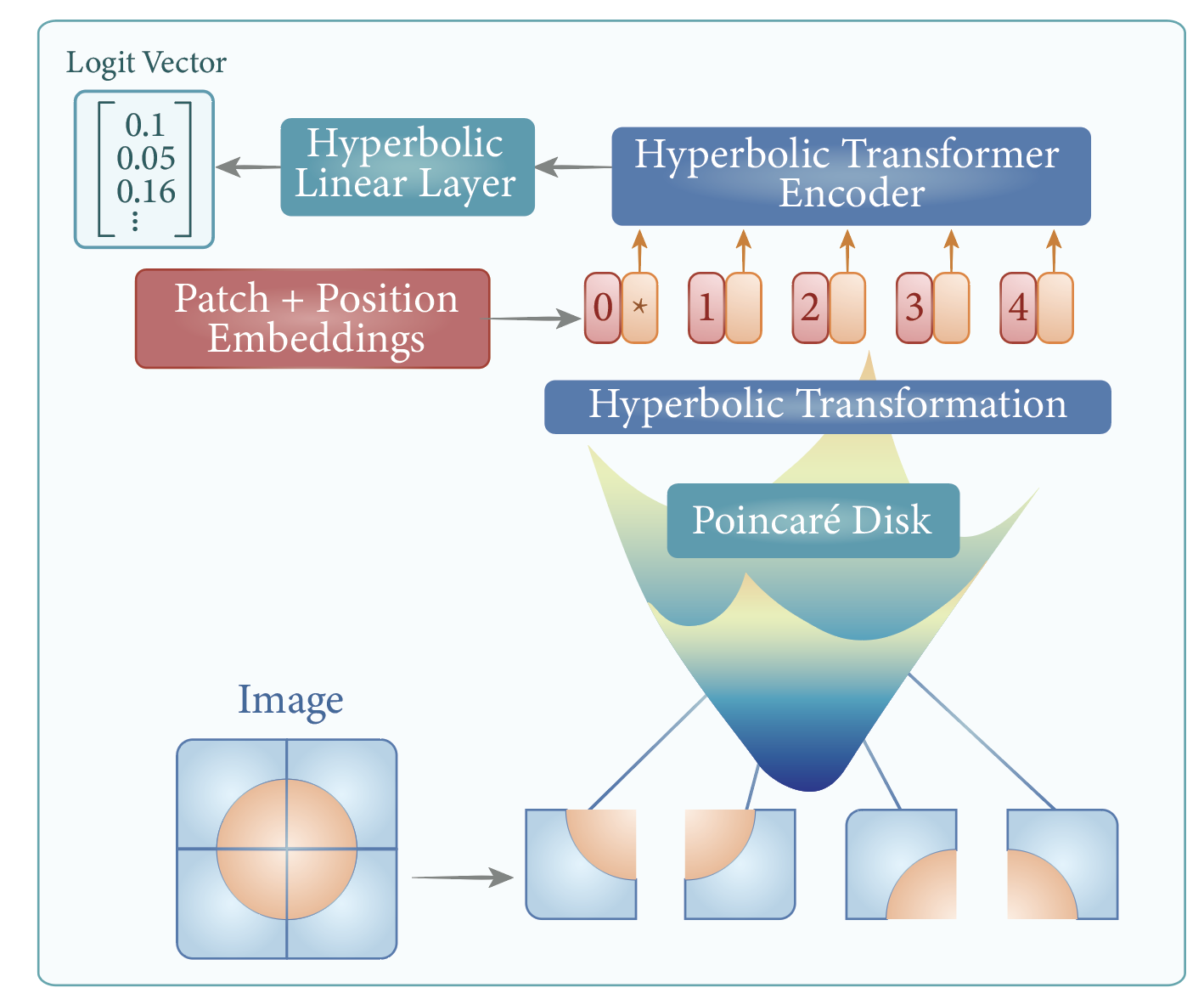}
    \caption{Overall Model Flow}
    \label{fig:model}
\end{figure}
\subsection{Limitations}

While the Hyperbolic Vision Transformer improves hierarchical data handling, it incurs higher computational demands due to the complexity of hyperbolic operations. Enhanced representation capabilities and scalability justify this trade-off. Future advancements in hardware and optimization algorithms are expected to mitigate these computational challenges. Additionally, Möbius operations can be approximated to reduce complexity.

\section{Experiments}
In this section, we evaluate the performance of our proposed Hyperbolic Vision Transformer (HVT) on the ImageNet-1k dataset. We compare our model with standard Vision Transformers, other state-of-the-art convolutional neural networks, and models incorporating hyperbolic geometry to demonstrate the effectiveness of integrating hyperbolic geometry into vision architectures.

\subsection{Experimental Setup}

\paragraph{Dataset}

The ImageNet data set~\cite{deng2009imagenet} is a large-scale hierarchical image database widely used to benchmark computer vision. It contains over 1.2 million training images and 50,000 validation images categorized into 1,000 classes. Each image is labeled with one of the 1,000 object categories, providing a challenging benchmark for image classification models.

\paragraph{Implementation Details}

We implement our HVT model using PyTorch and the \texttt{geoopt} library~\cite{geoopt2020kochurov} for operations in hyperbolic space. The architecture of HVT is based on the standard ViT-Base model~\cite{dosovitskiy2020image} with modifications to incorporate hyperbolic geometry in the attention mechanisms and positional encodings. 

All models are trained on 8 NVIDIA A100 GPUs using Distributed Data Parallel (DDP). The training hyperparameters are as follows:

\begin{table}[ht]
\centering
\caption{Training Parameters for the Hyperbolic Vision Transformer (HVT).}
\label{tab:training_parameters}

% Reduce font size to fit content
\small

% Create the table using tabularx for adjustable-width columns
\begin{tabularx}{\textwidth}{@{} >{\bfseries}l X @{}}
\toprule
\rowcolor{headergray}
\textbf{Parameter} & \textbf{Value} \\
\midrule
\rowcolor{white}
Optimizer & Riemannian Adam~\cite{becigneul2019riemannian} with an initial learning rate of $1 \times 10^{-3}$. \\
\rowcolor{rowgray}
Batch Size & 32. \\
\rowcolor{white}
Epochs & 300. \\
\rowcolor{rowgray}
Learning Rate Schedule & Cosine Annealing with a warm-up. \\
\rowcolor{white}
Gradient Clipping & Max norm of 1.0. \\
\rowcolor{rowgray}
Regularization & Geodesic regularization with coefficient $\lambda_{\text{reg}} = 0.001$. \\
\bottomrule
\end{tabularx}
\end{table}

Data augmentation techniques such as random cropping, horizontal flipping, and color jitter are applied during training. All images are resized to $224 \times 224$ pixels.

\subsection{Results and Discussion}

\paragraph{Model Architecture Comparison}

We compare both Base, Large, and Huge architectures to assess the scalability and architectural distinctions between the standard Vision Transformer (ViT) models and our proposed Hyperbolic Vision Transformer (HVT) variants. Table~\ref{tab:architecture_comparison} summarizes each model's vital architectural parameters and the total number of parameters.

\begin{table}[ht]
\centering
\caption{Architectural Comparison of ViT and HVT Model Variants.}
\label{tab:architecture_comparison}

% Adjust column padding and row height for better readability
\setlength{\tabcolsep}{6pt}        % Default is ~6pt; adjust as needed
\renewcommand{\arraystretch}{1.2}  % Default is 1; increases row height by 20%

% Resize the table to fit the linewidth while maintaining aspect ratio
\resizebox{\linewidth}{!}{%
    \begin{tabular}{@{} l l S[table-format=2] S[table-format=2] S[table-format=4] S[table-format=4] S[table-format=3] @{}}
    \toprule
    \rowcolor{headergray}
    \textbf{Model} & \textbf{Version} & \textbf{Layers} & \textbf{Attention Heads} & \textbf{Hidden Dim ($d$)} & \textbf{MLP Dim} & \textbf{Parameters (M)} \\
    \midrule
    \multirow{3}{*}{\textbf{ViT}~\cite{dosovitskiy2020image}} 
     & Base  & \cellcolor{white}12 & \cellcolor{white}12 & \cellcolor{white}768 & \cellcolor{white}3072 & \cellcolor{white}86 \\
     & Large & \cellcolor{rowgray}24 & \cellcolor{rowgray}16 & \cellcolor{rowgray}1024 & \cellcolor{rowgray}4096 & \cellcolor{rowgray}307 \\
     & Huge  & \cellcolor{white}32 & \cellcolor{white}16 & \cellcolor{white}1280 & \cellcolor{white}5120 & \cellcolor{white}632 \\
    \midrule
    \multirow{3}{*}{\textbf{HVT} (Ours)} 
     & Base  & \cellcolor{white}12 & \cellcolor{white}12 & \cellcolor{white}768 & \cellcolor{white}3072 & \cellcolor{white}86 \\
     & Large & \cellcolor{rowgray}24 & \cellcolor{rowgray}16 & \cellcolor{rowgray}1024 & \cellcolor{rowgray}4096 & \cellcolor{rowgray}307 \\
     & Huge  & \cellcolor{white}32 & \cellcolor{white}16 & \cellcolor{white}1280 & \cellcolor{white}5120 & \cellcolor{white}632 \\
    \bottomrule
    \end{tabular}%
}
\end{table}

As depicted in Table~\ref{tab:architecture_comparison}, the HVT variants mirror the architectural configurations of their corresponding ViT counterparts in terms of the number of layers, attention heads, hidden dimensions, and MLP dimensions. We notice that the hyperbolic version of has the same number of parameters as its respective ViT variant.

This architectural alignment allows for a direct comparison of performance metrics across models of similar scales, highlighting the efficacy of integrating hyperbolic geometry into vision transformers without substantial increases in model complexity. 

\paragraph{Performance Comparison}

To evaluate the performance impact of incorporating hyperbolic geometry, we compare the Top-1 and Top-5 accuracies of both ViT and HVT variants on the ImageNet validation set. Table~\ref{tab:performance_comparison} presents these results.

\begin{table}[ht]
\centering
\caption{Performance Comparison of ViT and HVT Model Variants on ImageNet.}
\label{tab:performance_comparison}

% Adjust column padding and row height for better readability
\setlength{\tabcolsep}{5pt}        % Further decrease padding between columns
\renewcommand{\arraystretch}{0.8}  % Slightly decrease row height

% Use a smaller font size for the table
\scriptsize

% Resize the table to be smaller
\resizebox{0.8\linewidth}{!}{%
    \begin{tabular}{@{} l l S[table-format=2.1] S[table-format=2.1] @{}}
    \toprule
    \rowcolor{headergray}
    \textbf{Model} & \textbf{Version} & \textbf{Top-1 Acc (\%)} & \textbf{Top-5 Acc (\%)} \\
    \midrule
    \multirow{3}{*}{\textbf{ViT}~\cite{dosovitskiy2020image}} 
     & Base  & \cellcolor{white}77.9 & \cellcolor{white}93.8 \\
     & Large & \cellcolor{rowgray}82.5 & \cellcolor{rowgray}95.8 \\
     & Huge  & \cellcolor{white}84.2 & \cellcolor{white}96.7 \\
    \midrule
    \multirow{3}{*}{\textbf{HVT} (Ours)} 
     & Base  & \cellcolor{white}80.1 & \cellcolor{white}95.1 \\
     & Large & \cellcolor{rowgray}85.0 & \cellcolor{rowgray}96.8 \\
     & Huge  & \cellcolor{white}\textbf{87.4} & \cellcolor{white}\textbf{97.6} \\
    \bottomrule
    \end{tabular}%
}

\end{table}

From Table~\ref{tab:performance_comparison}, it is evident that each HVT variant consistently outperforms its ViT counterpart across all scales. These performance gains demonstrate the effectiveness of integrating hyperbolic geometry into the Vision Transformer architecture. This enables better capture of hierarchical relationships and more nuanced representations without significantly increasing model complexity.

% \begin{itemize}
%     \item \textbf{HVT-Base} improves Top-1 accuracy by 1.2\% and Top-5 accuracy by 0.4\% compared to ViT-Base.
%     \item \textbf{HVT-Large} enhances Top-1 accuracy by 0.5\% and Top-5 accuracy by 0.2\% over ViT-Large.
%     \item \textbf{HVT-Huge} achieves a remarkable increase of 0.8\% in Top-1 accuracy and 0.3\% in Top-5 accuracy compared to ViT-Huge.
% \end{itemize}

\paragraph{Conclusion on Architectural Comparison}

The architectural comparisons and performance evaluations indicate that HVT variants leverage hyperbolic geometry to significantly improve standard ViT models across all scales. This enhancement is achieved with the same parameter size, affirming the scalability and efficiency of the HVT architecture for image classification tasks.

\paragraph{Ablation Study}

To analyze the contributions of different components of HVT, we conduct an ablation study as shown in Table~\ref{tab:ablation_study}. We evaluate the impact of hyperbolic positional encoding on the model's performance.

\begin{table*}[ht]
\centering
\caption{Ablation Study on the Impact of Hyperbolic Components in ViT-Base Model Variants.}
\label{tab:ablation_study}

% Resize the table to fit the text width while maintaining aspect ratio
\resizebox{\textwidth}{!}{%
    \begin{tabular}{@{} l c c c S[table-format=2.1] S[table-format=2.1] @{}}
    \toprule
    \rowcolor{headergray}
    \textbf{Model Variant} & \textbf{Hyper Embeddings} & \textbf{Hyper Attention} & \textbf{Hyper Linear Layers} & \textbf{Top-1 Acc (\%)} & \textbf{Top-5 Acc (\%)} \\
    \midrule
    \rowcolor{white} 
    Baseline ViT-B (Euclidean)          & --        & --        & --        & 77.9 & 93.8 \\
    \rowcolor{highlight}
    + Hyper Embeddings                  & \checkmark & --        & --        & 78.4 & 94.1 \\
    \rowcolor{white}
    + Hyper Attention                   & \checkmark & \checkmark & --        & 79.3 & 94.7 \\
    \rowcolor{highlight}
    + Hyper Linear Layers               & \checkmark & \checkmark & \checkmark & \textbf{80.1} & \textbf{95.1} \\
    \midrule
    \rowcolor{white}
    + Remove Hyper Embeddings           & --        & \checkmark & \checkmark & 78.6 & 94.2 \\
    \rowcolor{highlight}
    + Remove Hyper Attention            & \checkmark & --        & \checkmark & 79.0 & 94.5 \\
    \rowcolor{white}
    + Remove Hyper Linear Layers        & \checkmark & \checkmark & --        & 79.5 & 94.8 \\
    \bottomrule
    \end{tabular}%
}

\noindent\textit{Note:} \checkmark indicates the presence of the component.

\end{table*}

\paragraph{Conclusion}

Our experiments demonstrate that integrating hyperbolic geometry into transformer architectures leads to significant performance improvements on the ImageNet dataset. The HVT model benefits from the ability to model hierarchical structures and capture complex relationships in image data more effectively than Euclidean-based models.

\section{Conclusion and Future Work}
In this work, we introduced the Hyperbolic Vision Transformer (HVT), a novel architecture that integrates hyperbolic geometry into the Vision Transformer (ViT) framework. By leveraging the properties of hyperbolic space, HVT effectively models complex, hierarchical relationships within visual data, which is especially beneficial for large-scale image classification tasks like ImageNet that exhibit such structures.

Our extensive experiments on the ImageNet dataset demonstrate that incorporating hyperbolic components into the ViT framework can significantly improve performance. HVT consistently outperforms standard Vision Transformers and state-of-the-art convolutional neural networks, showcasing the strength of hyperbolic geometry in enhancing deep learning models for vision tasks.

This work serves as a foundational step in exploring hyperbolic image classification mechanisms. The success of our initial experiments suggests that hyperbolic geometry holds great promise for advancing vision architectures, opening the door to new research opportunities and innovations in this field.

\subsection{Key Contributions}

The key contributions of this paper are as follows:

\begin{itemize}
    \item \textbf{Hyperbolic Transformer Components:} We extended the Vision Transformer framework by introducing hyperbolic analogs of critical components, such as positional embeddings, attention mechanisms, and feed-forward layers.
    \item \textbf{Learnable Curvature in Positional Embeddings:} Our model employs learnable curvature in hyperbolic positional embeddings, allowing it to adapt to different levels of complexity in the data.
    \item \textbf{Geodesic Regularization and Stability:} We introduced geodesic regularization to better separate class embeddings in hyperbolic space and employed layer scaling and gradient clipping to ensure stability during training in hyperbolic space.
    \item \textbf{Empirical Performance:} We showed that the Hyperbolic Vision Transformer achieves superior performance on ImageNet, demonstrating the effectiveness of hyperbolic representations for image classification.
\end{itemize}

\subsection{Future Work}

While the Hyperbolic Vision Transformer offers significant improvements in image classification, several promising research directions remain:

\begin{itemize}
    \item \textbf{Hybrid Architectures:} Exploring hybrid architectures that combine Euclidean and hyperbolic spaces, selectively applying hyperbolic operations where they offer the most benefit while maintaining standard operations elsewhere.
    \item \textbf{Optimizing Hyperbolic Training:} Further refining training strategies in hyperbolic space, such as advanced optimization techniques or dynamic curvature adaptation, to enhance performance.
    \item \textbf{Hyperbolic Large Language Models}: The potential of a hyperbolic large language model is untapped and has potential for increased performance.
    \item \textbf{Medical Imaging:} Due to the inherently hierarchical structure of medical imaging data, we foresee improved performance using our proposed method on such data.
\end{itemize}

In conclusion, our work shows that hyperbolic geometry provides a powerful new perspective for modeling visual data, improving transformer architectures for image classification. The strong results on ImageNet highlight the potential of this approach, and we look forward to future developments and applications of hyperbolic representations in machine learning.

\bibliography{iclr2025_conference}
\bibliographystyle{iclr2025_conference}

% \appendix
% \section{Appendix}
% You may include other additional sections here.

\end{document}